\documentclass{SPAICE}

\def\authorEmail{toon.vinck@magics.tech}
\def\AuthorShort{T. Vinck et al.}

\author[1,2,3]{Toon Vinck\thanks{Corresponding author. E-Mail: \authorEmail}}
\author[2]{Naïn Jonckers}
\author[1]{Jaro De Roose}
\author[2]{Jeffrey Prinzie}
\author[3,4]{Peter Karsmakers}

\affil[1]{Magics Technologies, Geel, Belgium}
\affil[2]{ESAT-ADVISE, KU Leuven, Geel, Belgium}
\affil[3]{Dept. of Computer Science, Leuven.AI, KU Leuven, Leuven, Belgium}
\affil[4]{MPRO, Flanders Make, Leuven, Belgium}

\title{Exploring the Trade-Off Between Structured Pruning and Fault Tolerance in Deep Neural Networks for Space Applications}

\begin{document}

\maketitle

\begin{abstract}
Deep Neural Networks (DNNs) inherently exhibit a degree of robustness to bit-level faults due to their distributed representation of information. As a model increases in width, this information becomes more dispersed, theoretically reducing the impact of any single bit fault. In this paper, we empirically investigate the relationship between model width and robustness to Single Event Upsets (SEUs). We conduct a comprehensive experiment in which baseline models undergo iterative structured pruning to reduce their width while preserving task performance as much as possible. At each pruning stage, we run a targeted fault-injection campaign to evaluate the model's performance under simulated bit-flip scenarios. Our results show that, although structured pruning increases per-inference sensitivity to faults by reducing redundancy, this effect is effectively counterbalanced by shorter execution time, which lowers the probability of encountering an SEU. These findings suggest that structured pruning can yield significant energy and latency savings without compromising overall reliability, providing useful guidance for designing robust AI systems for space applications.
\end{abstract}

\section{Introduction}
The rapid advancement of Artificial Intelligence (AI) algorithms, particularly Deep Neural Networks (DNNs), has led to remarkable performance improvements. This progress has fueled growing interest in deploying these algorithms in space applications. However, the space environment presents unique hazards, such as ionizing radiation, which can induce computational errors known as Single Event Upsets (SEUs). While state-of-the-art radiation-hardened AI processors are still not widely available, DNNs inherently possess a degree of redundancy due to their parallel and distributed processing architecture \cite{rech2024artificial}. Errors in the computation of individual neurons may significantly affect that neuron's output. However, because each neuron typically contributes only partially to the inputs of subsequent layers, the overall impact on network performance is often limited. Furthermore, this structural redundancy, combined with the masking effects of non-linear activation functions and pooling layers, means that isolated bit faults are frequently mitigated before they can influence the final classification \cite{li2017understanding, malekzadeh2021impact}.\par
However, there is an inherent trade-off: this redundancy implies that the network is larger than strictly necessary, consuming valuable memory and energy resources. This has driven the development of pruning techniques to compress models, but it raises a critical question: does removing this redundancy compromise robustness?\par
Previous research has investigated the relationship between pruning and fault robustness. Guillemé et al. \cite{guilleme2025fault} observed that removing individual connections (unstructured pruning) generally makes models more sensitive to single faults per inference. In contrast, Goldstein et al. \cite{goldstein2020reliability} accounted for the fact that smaller models perform fewer computations. Because the likelihood of a particle strike is proportional to the duration of an operation, reduced execution time decreases the probability of encountering an SEU during inference. They therefore concluded that network pruning can improve the reliability. Nevertheless, these studies focused on unstructured pruning, which does not alter the width of the network but instead increases the network's sparsity. Moreover, many simple edge accelerators do not exploit unstructured sparsity, limiting the practical speedups and energy savings achievable with unstructured pruning. Finally, these studies did not consider faults in high-bit-width accumulator registers found in quantized DNN accelerators, which can be a significant source of faults due to their large dynamic range \cite{vinck2024understanding}.\par
To address this gap, we investigate the relationship between global structured pruning, which removes entire channels from a DNN, and fault tolerance. We evaluate fault tolerance via Fault Injection (FI) simulation in a simple DNN accelerator architecture that includes a high-bit-width accumulator.\par
The remainder of this paper is organized as follows. \cref{sec:experiment} describes the experiment, including the FI framework and the methodology used to create models of varying width. \cref{sec:results} presents the results of the FI campaign on pruned models, followed by a discussion in \cref{sec:discussion}. Finally, \cref{sec:conclusion} concludes the paper and outlines potential directions for future research.

\section{Experiment}
\label{sec:experiment}

\subsection{Fault Injection Framework}


DNN accelerators commonly consist of a set of Processing Elements (PEs) that perform Multiply--Accumulate (MAC) operations, which are the core computations in DNN inference. To execute these computations efficiently, PEs often use quantized arithmetic. In addition, parameters and/or activations are reused from local registers rather than fetched from off-chip memory for every operation, improving performance and reducing power consumption.
For example, the original Google TPU \cite{jouppi2017datacenter} employs a systolic-array architecture in which weights are preloaded into the PEs. Activation data flows through the array, and partial sums are passed to adjacent PEs. Similarly, Eyeriss \cite{chen2016eyeriss} uses a row-stationary dataflow: filter rows of 2D convolutional layers stream through the array, while input feature-map rows are kept stationary in the PEs to maximize local data reuse.


\begin{figure}[t]
    \centering
        \includegraphics[width=\columnwidth]{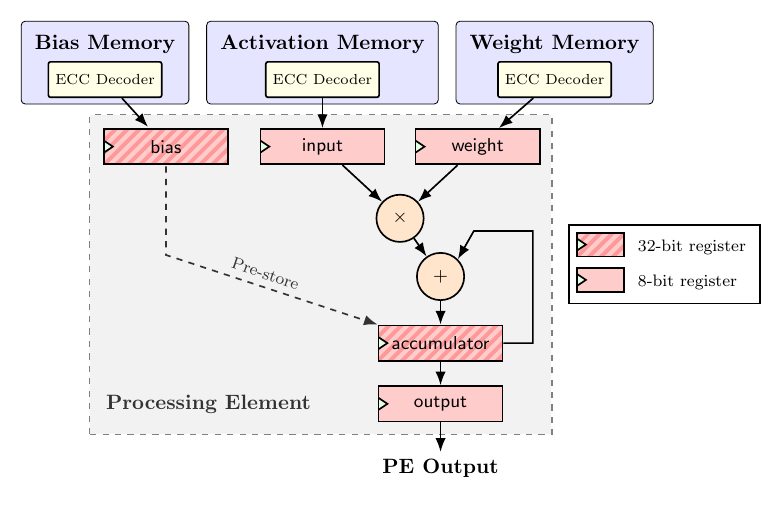}
    \caption{Abstract representation of a simple Processing Element.}
    \label{fig:simple_pe}
\end{figure}

Because each accelerator architecture has its own dataflow and reuse strategy, SEUs can have complex and accelerator-dependent effects on DNN inference. To simplify the analysis, we implemented a FI framework that simulates the dataflow of a single, simple PE, as shown in \cref{fig:simple_pe}.
Furthermore, because this paper targets space applications, we assume that the main memories in this dataflow are protected from SEUs by hardware Error Correction Codes (ECC). This protection is important because, without it, faults could accumulate and degrade DNN performance over time. In contrast to PEs, memory elements can often be protected efficiently, while protecting PEs typically incurs substantial area and power overhead. Therefore, we exclude main memory from the FI tool and focus on the vulnerability of the PEs themselves.


The FI tool is built on top of the PyTorch \cite{NEURIPS2019_bdbca288} quantization package Brevitas \cite{brevitas} and uses fake-quantized DNNs following the guidance of Nagel \cite{nagel2021white}. As shown in \cref{fig:simple_pe}, in each DNN layer, 8-bit inputs and weights are multiplied and accumulated in a 32-bit accumulator register, together with a preloaded 32-bit bias. The final accumulator value is then rounded to an 8-bit output using a power-of-two scale factor.

The FI process consists of three main steps:
\begin{enumerate}
    \item \textbf{Calibration}: The tool analyzes the DNN inference to map all bits in the five modules (input, weight, bias, accumulator and output) to a global address space.
    \item \textbf{Selection}: Before each inference, one bit in the entire DNN inference trace is selected randomly.
    \item \textbf{Injection}: During inference, the selected bit is flipped at the targeted DNN state using PyTorch module hooks \cite{pytorch_module_hooks}.
\end{enumerate}

We inject only a single bit flip because the probability of multiple independent SEUs occurring during a single inference (a millisecond-scale time window) is very low in typical space environments \cite{jonckers2024single}. For simplicity, we also ignore the possibility that a single ionizing particle can cause multiple bit flips (multi-bit upsets).

\subsection{Methodology}

\begin{figure}[t]
    \centering
    \includegraphics[width=\columnwidth]{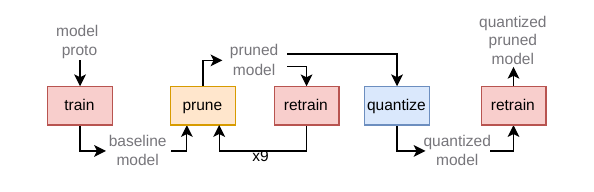}
    \caption{Flowchart of the iterative pruning and quantization methodology.}
    \label{fig:deployment_flow}
\end{figure}

To systematically evaluate the relationship between global structured pruning and robustness, we use the experimental pipeline shown in \cref{fig:deployment_flow}.

\begin{itemize}
    \item \textbf{Baseline training}: We train the baseline models to convergence on the target dataset.
    \item \textbf{Iterative global structured pruning}: To create models of varying width, we apply iterative global structured pruning using the \texttt{torch-pruning} library \cite{fang2023depgraph}. We use the grouped first-order Taylor expansion of the loss function \cite{molchanov2019importance} as a global importance metric and prune the least important channels. The final layer is kept unpruned to preserve the output dimensionality. The process is performed in 10 iterations, pruning 14.9\% of the remaining output channels at each step, which yields a final model that retains 20\% of the original channels. After each pruning step, the model is retrained from the remaining (unpruned) channels to recover accuracy.
    \item \textbf{Quantization}: The resulting models undergo Quantization-Aware Training (QAT) to adapt them to 8-bit integer arithmetic.
    \item \textbf{Fault Injection}: Finally, we run a FI campaign on each model variant. We measure the performance drop under fault conditions to quantify robustness as a function of model width.
\end{itemize}

We conduct two experiments using ResNet-18 \cite{he2016deep} and MobileNetV2 \cite{sandler2018mobilenetv2} on the EuroSAT dataset \cite{helber2018introducing}. The dataset is randomly split into 80\% training, 10\% validation, and 10\% test sets.

To improve model generalization, we apply standard data augmentation to the training set, including random cropping with 4-pixel padding (applied before cropping), horizontal and vertical flips, and discrete random rotations of $0^{\circ}, 90^{\circ}, 180^{\circ} \text{or } 270^{\circ}$. All images are resized to $64 \times 64$ pixels and normalized to ensure a consistent input distribution.

Detailed training hyperparameters for the baseline training, pruning retraining (per iteration), and QAT phases are provided in \cref{tab:training_recipes}. Both models are optimized using Stochastic Gradient Descent (SGD) with a momentum of 0.9, a weight decay of $1 \times 10^{-4}$, and a batch size of 128. The learning rate is reduced by a factor of 0.1 at the listed schedule milestones (specific epochs listed in \cref{tab:training_recipes}). In each training phase, we use a callback to save the best model based on validation accuracy. For the robustness analysis, we inject 102{,}600 faults into each model at every pruning iteration.\par

\begin{table}[h]
\centering
\begin{tabular}{lcc}
\toprule
\textbf{Parameter} & \textbf{ResNet-18} & \textbf{MobileNetV2} \\
\midrule
Dataset & \multicolumn{2}{c}{EuroSAT \cite{helber2018introducing}} \\
Optimizer & \multicolumn{2}{c}{SGD} \\
Batch Size & \multicolumn{2}{c}{128} \\
Momentum & \multicolumn{2}{c}{0.9} \\
Weight Decay & \multicolumn{2}{c}{$1 \times 10^{-4}$} \\
LR Schedule $\gamma$ & \multicolumn{2}{c}{0.1} \\
\midrule
\multicolumn{3}{l}{\textit{Baseline Training}} \\
Base LR & \multicolumn{2}{c}{0.1} \\
Epochs & \multicolumn{2}{c}{140} \\
Milestones & [80, 100, 120] & [80, 120] \\
\midrule
\multicolumn{3}{l}{\textit{Pruning (per iteration)}} \\
Base LR & \multicolumn{2}{c}{0.1} \\
Epochs & \multicolumn{2}{c}{100} \\
Milestones & [40, 60, 80] & [40, 80] \\
\midrule
\multicolumn{3}{l}{\textit{Quantization (QAT)}} \\
Base LR & $1 \times 10^{-4}$ & $1 \times 10^{-3}$ \\
Epochs & \multicolumn{2}{c}{30} \\
\bottomrule
\end{tabular}
\caption{Hyperparameters for Training, Pruning, and Quantization.}
\label{tab:training_recipes}
\end{table}

\section{Results}

\label{sec:results}
\begin{figure}[b!]
    \centering
    \includegraphics[width=\columnwidth]{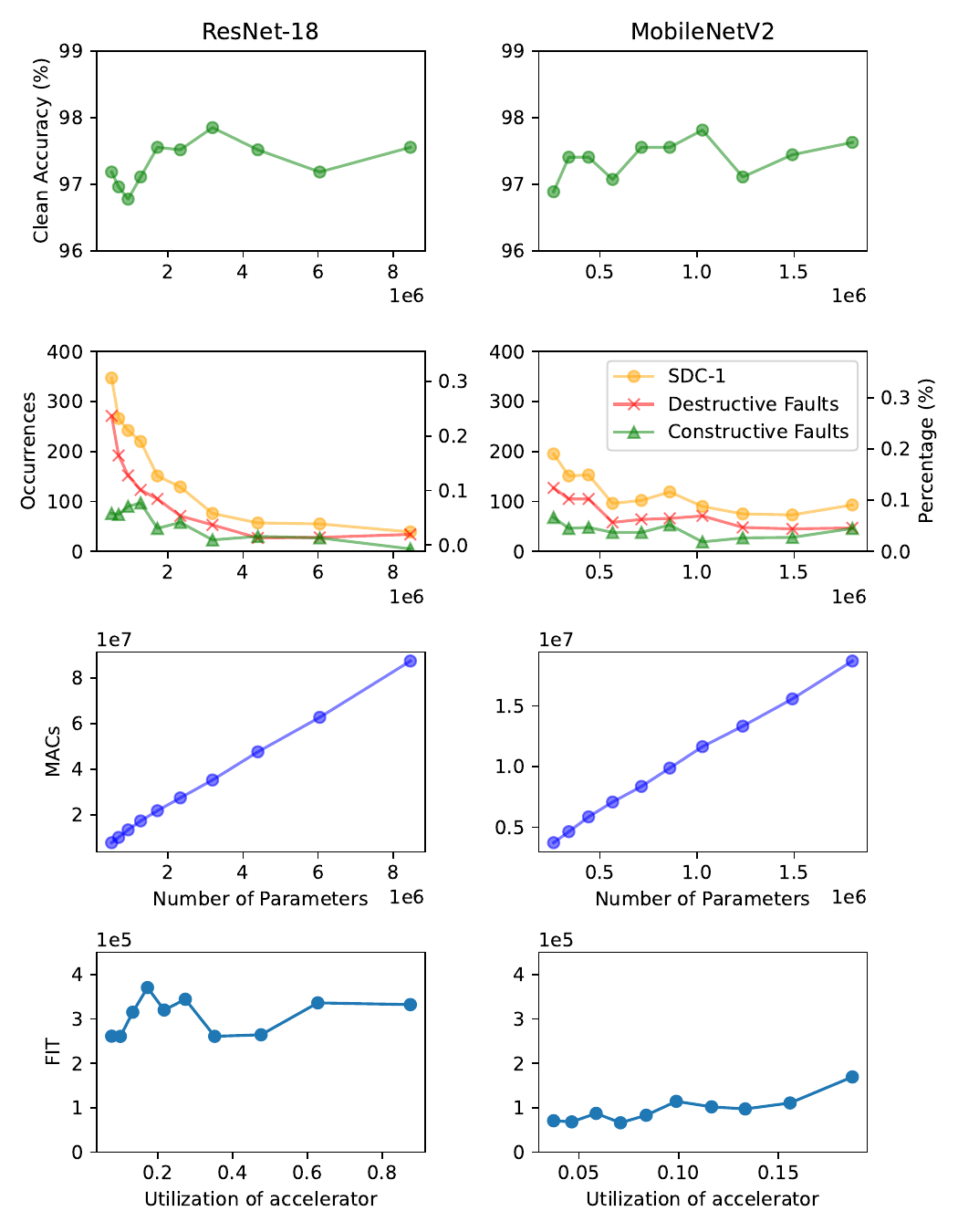}
    \caption{Results of the FI experiments.}
    \label{fig:fi_results}
\end{figure}

The results of the FI experiments on the pruned ResNet-18 and MobileNetV2 models are shown in \cref{fig:fi_results}. The plots are organized with ResNet-18 in the left column and MobileNetV2 in the right column.

The first row reports the reference accuracy (without FI) for each pruning stage. Due to the stochastic nature of DNN training, final test accuracy can vary slightly across runs. The proposed training-pruning procedure is designed to minimize this variability and maintain consistent performance across iterations. Nevertheless, a difference of up to 1\% may still occur between the best- and worst-performing models.

The second row presents the Silent Data Corruption (SDC) rate using the SDC-1 metric. This metric measures how often the top-1 predicted class under FI differs from the fault-free execution \cite{li2017understanding}. It includes both destructive faults (a correct classification becomes incorrect) and constructive faults (an injected fault corrects an otherwise incorrect classification).\par

The third row shows the total number of MAC operations required by each model. Assuming each MAC occupies a PE for a fixed duration, the operation count serves as a proxy for the probability of an SEU occurring during inference. Pruning reduces the number of MAC operations, with the most aggressive pruning level requiring less than 20\% of the original budget. Consequently, the model's exposure to potential SEUs drops proportionally to the same level.\par

Finally, the fourth row combines these metrics into a calculated Failure In Time (FIT) rate, defined as the number of failures per $10^9$ device-hours \cite{rech2024artificial}. We assume a hypothetical setup in which an accelerator running at 1~GMAC/s experiences 1~SEU/h at full utilization. For a target throughput of 10 inferences per second, the reduced operation count of pruned models lowers device utilization and therefore proportionally decreases the probability of encountering a vulnerable SEU.

For both architectures, the SDC-1 rate increases as model size is reduced. Moreover, the proportion of destructive faults appears to grow with more aggressive pruning. This results in a larger gap between clean and FI accuracy, indicating that inherent fault tolerance diminishes as network width decreases.

The hypothetical FIT rate combines two opposing effects: decreased redundancy in pruned models and a reduced probability of encountering an SEU due to fewer operations. For ResNet-18, these factors largely cancel out, yielding a relatively stable FIT rate. For MobileNetV2, we observe a slight decrease in FIT as the model becomes smaller, although the reduction appears limited.

Regarding the overall performance impact, the fraction of destructive faults remains small (<0.3\%), suggesting that although pruning increases sensitivity, the overall performance degradation is limited. In addition, MobileNetV2 exhibits substantially higher inherent fault tolerance than ResNet-18, with a calculated FIT rate that is 2-3x lower. This can be attributed to MobileNetV2's inverted residual architecture \cite{sandler2018mobilenetv2}. Although the model is computationally lighter due to depth-wise separable convolutions, the number of ``lightweight'' channels is as high as in ResNet-18. Consequently, information remains distributed across a large number of channels, preserving redundancy.

\section{Discussion}
\label{sec:discussion}

The observed increase in SDC-1 probability per inference for pruned models supports the hypothesis that model width contributes substantially to fault tolerance. In wider networks, the information required for classification is distributed across more channels. As a result, corruption in a single neuron's computation has a smaller relative effect, which helps masking the error and effectively providing inherent redundancy. Structured pruning systematically removes channels and therefore removes this internal redundancy. The remaining neurons become more critical to the network output, such that a single bit flip has a larger proportional impact, increasing the likelihood of destructive faults.

From a mission-level perspective, however, the reliability implications are more nuanced. The relative stability of the hypothetical FIT rate indicates that increased per-inference sensitivity is largely counterbalanced by the reduced computational footprint. Because pruned models require fewer operations, the time window during which an SEU can occur is shorter. This yields a useful design insight for space-bound systems: structured pruning can provide substantial savings in memory, energy, and latency without substantially increasing the system-level failure rate. In other words, although the model becomes algorithmically more fragile, the system can remain operationally robust due to reduced radiation exposure time.


Nevertheless, one might question the need for strict fault tolerance when DNNs already have imperfect accuracy. If fewer than 0.3\% of injected faults cause an incorrect classification, while a fault-free model already has a baseline error rate---which depends on the application but is approximately 2--3\% in this study---the added risk may appear low. In safety-critical space applications, however, the implications of faults can be profound. Unlike epistemic uncertainty, which arises from limited model knowledge and typically manifests on out-of-distribution inputs, radiation-induced SDCs introduce stochastic failures that are unrelated to the data distribution. While epistemic uncertainty reflects model limitations and aleatoric uncertainty captures intrinsic input noise \cite{hullermeier2021aleatoric}, radiation faults perturb the network's internal computations. As a consequence, SDCs can cause catastrophic, high-confidence misclassifications even for trivial inputs that would otherwise be correctly classified under fault-free conditions \cite{rech2024artificial}. Therefore, even if the average accuracy degradation due to radiation is small, minimizing the probability of these unpredictable, hardware-induced failures remains essential for safety certification and mission assurance.


\section{Conclusion}
\label{sec:conclusion}
Our experiments show that structured pruning makes models less tolerant to faults because it reduces redundancy. However, this increased per-inference sensitivity is mitigated by the lower computational load of pruned models, which reduces the probability of encountering an SEU during inference.

We therefore conclude that optimizing for efficiency does not compromise reliability and suggest that, when designing space-borne AI systems, priority should be given to energy- and time-efficient models. In addition, the inherent fault tolerance of DNNs can be leveraged to simplify accelerator design for space applications by relaxing strict radiation-hardening requirements in the hardware.

Although the observed number of SDC events is low, verifying that their impact on safety-critical applications is negligible remains a key concern. Future work should extend this analysis by assessing fault impact at the mission level and distinguishing between benign anomalies and critical failures.
 

\begin{acknowledgments}
This work was funded through a VLAIO Baekeland mandate. The authors further acknowledge the support and collaboration framework provided by the VLAIO MEDLI COOCK+ project (HBC.2024.0597).
\end{acknowledgments}

\printbibliography
\addcontentsline{toc}{section}{References}

\end{document}